\documentclass{article}
\usepackage{ijcai25}
\usepackage{times}
\usepackage{soul}
\usepackage{url}
\usepackage[hidelinks]{hyperref}
\usepackage[utf8]{inputenc}
\usepackage[small]{caption}
\usepackage{graphicx}
\usepackage{amsmath}
\usepackage{amsthm}
\usepackage{booktabs}
\usepackage{algorithm}
\usepackage{algorithmic}
\usepackage[switch]{lineno}
\usepackage{multirow}

\usepackage[accsupp]{axessibility}  

\usepackage{orcidlink}
\usepackage{soul}

\title{SyncAnimation: A Real-Time End-to-End Framework for Audio-Driven Human Pose and Talking Head Animation} 

\author{Yujian Liu$^{1,2}$ \and
Shidang Xu$^{2,}$\thanks{Equal contribution.} \and
Jing Guo  $^{1,3}$ \and
Dingbin Wang  $^{1,4}$ \and
Zairan Wang$^1$ \and \\
Xianfeng Tan$^1$ \and 
Xiaoli Liu$^{1,}$\thanks{Corresponding author.} \\
\affiliations
$^1$AiShiWeiLai AI Research, Beijing, China \\
$^2$South China University of Technology, Guangzhou, China \\
$^3$Beijing Institute of Technology, Beijing, China\\
$^4$Beijing University of Posts and Telecommunications, Beijing, China\\
}

\begin{document}

\twocolumn[{
\renewcommand\twocolumn[1][]{#1}
\maketitle
\begin{center}
    \captionsetup{type=figure}
    \includegraphics[width=0.99\textwidth]{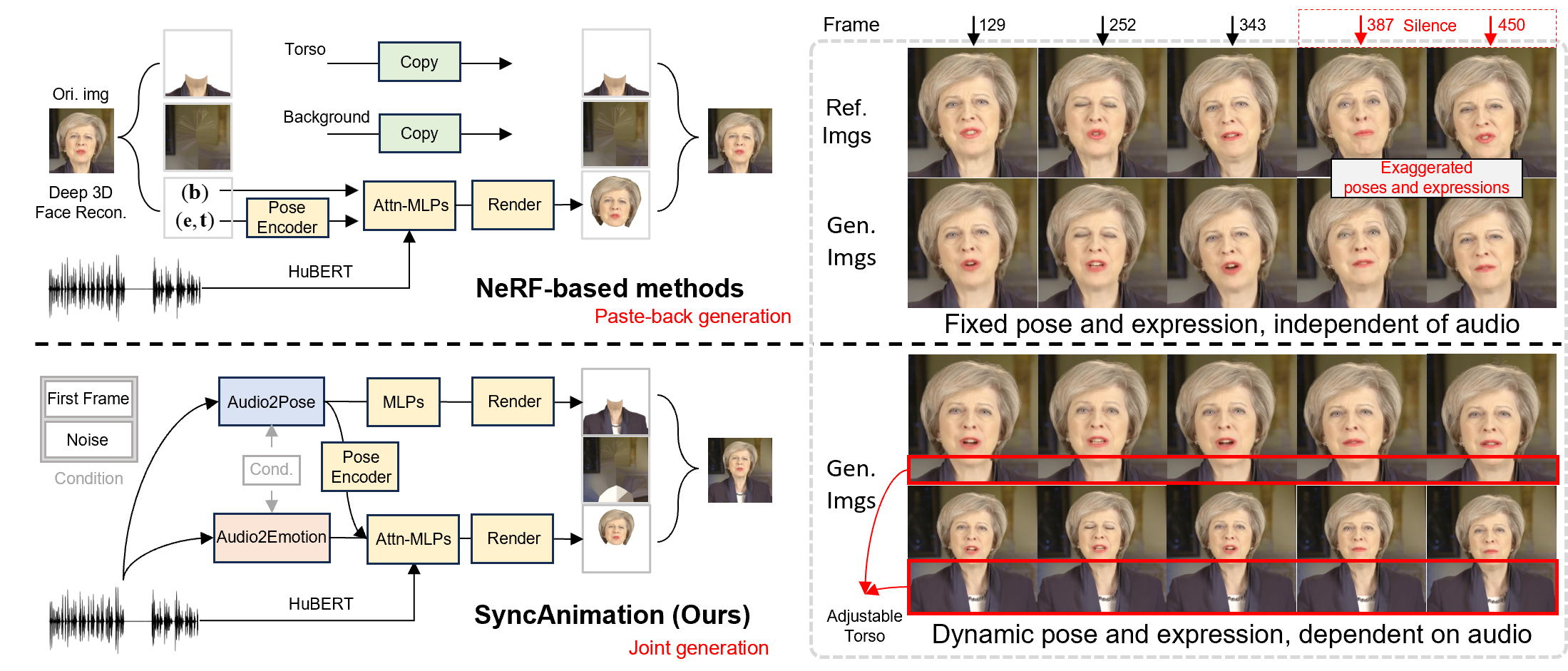}
    \captionof{figure}{SyncAnimation is the first NeRF-based jointly generative approach that utilizes audio-driven generation to create expressions and an adjustable upper body (left). SyncAnimation requires only audio and monocular, or even noise, to render highly detailed identity information, along with realistic and dynamic facial and upper-body changes, while maintaining audio consistency (right).}
    \label{fig:metadreamer_showcase}
\end{center}
\vspace{5mm}
}]

\renewcommand{\thefootnote}{\fnsymbol{footnote}} 
\footnotetext[1]{Equal contribution.} 
\footnotetext[2]{Corresponding author.} 

\begin{abstract}
Generating talking avatar driven by audio remains a significant challenge. Existing methods typically require high computational costs and often lack sufficient facial detail and realism, making them unsuitable for applications that demand high real-time performance and visual quality. Additionally, while some methods can synchronize lip movement, they still face issues with consistency between facial expressions and upper body movement, particularly during silent periods. In this paper, we introduce SyncAnimation, the first NeRF-based method that achieves audio-driven, stable, and real-time generation of speaking avatar by combining generalized audio-to-pose matching and audio-to-expression synchronization. By integrating AudioPose Syncer and AudioEmotion Syncer, SyncAnimation achieves high-precision poses and expression generation, progressively producing audio-synchronized upper body, head, and lip shapes. Furthermore, the High-Synchronization Human Renderer ensures seamless integration of the head and upper body, and achieves audio-sync lip. The project page can be found at \href{https://syncanimation.github.io/}{https://syncanimation.github.io/}

\end{abstract}

\section{Introduction}
\label{sec:intro}
In recent years, audio-visual synthesis techniques have garnered significant attention, with audio-driven realistic avatar generation emerging as a key research focus. 
Over the past few years, many researchers have employed GAN-based or SD-based deep generative models to tackle this task~\cite{prajwal2020lip,zhang2023dinet,zhong2023identity,zhang2023sadtalker,xu2024hallo,wang2024v,chen2024echomimic,xie2024x,tan2025edtalk}. Among them, SD-based models, leveraging model parameters and large-scale datasets, can generate fully animated avatars from a single reference image. However, their reliance on large-scale and diverse individual datasets, coupled with overwhelming computational and time costs, limits their applicability in real-time scenarios such as live streaming or video conferencing, where high fidelity and real-time rendering are essential ~\cite{guo2024i2v,hu2024animate}.

Recently, Neural Radiance Fields (NeRF) have been applied to audio-driven talking avatars \cite{mildenhall2021nerf,guo2021ad,li2023efficient,ye2023geneface,kim2024nerffacespeech,peng2024synctalk,ye2023geneface++}. By associating NeRF with audio either end-to-end or through intermediate representations, these methods enable the reconstruction of personalized talking avatars with impressive synthesis quality and inference speed. However, existing approaches, such as ER-NeRF~\cite{li2023efficient} and Synctalk~\cite{peng2024synctalk}, focus primarily on achieving precise synchronization between lip movement and audio, given their strong correlation. Nevertheless, they have yet to tackle the mismatch between audio and head poses, as well as the challenging association between audio and facial expressions~\cite{zhang2023sadtalker}, ultimately reducing the overall realism of the generated avatar.

In this paper, we highlight the importance of generating realistic talking avatars driven by audio, focusing on identity consistency and facial detail preservation, while faceing three critical challenges that require further attention, shown in Fig.~\ref{fig:metadreamer_showcase}: 
(1). Pose inconsistency with audio: Generate identical and fixed poses across different inferred audios (derived from original video frames), possibly even exhibiting exaggerated head movement in silent segments. (2). Expression inconsistency with audio: Insufficient attention is given to facial animation elements beyond lip-syncing, such as eyebrow movement and blinking, which are crucial to conveying natural expressions and emotional depth, resulting in unnatural and stiff animations. (3). Loss of paste-back ability in audio-driven pose method: Only the head is generated, and the edges of the body are generated with changes, causing the torso to be displaced and therefore cannot be attached back to the origin torso. As mentioned above, audio-driven pose generation methods cannot achieve upper-body generation, and the lack of audio-driven expression generation results in inconsistent expressions. Thus, a fully generative NeRF-based approach with strong audio correlation is essential for achieving realistic talking avatars.

To address these critical challenges, we propose SyncAnimation, a NeRF-based framework focused on audio-driven rendering of upper body and head . This framework integrates three core modules: the AudioPose Syncer and AudioEmotion Syncer, which enable stable, precise, and controllable mappings from audio to head poses and facial expressions, and the High-Synchronization Human Renderer, which ensures seamless integration of head motion and upper-body movement without post-processing. Together, these modules form a unified solution for generating dynamic, expressive, and highly audio-synchronized avatars. In summary, the main contributions of our work are as follows:
\begin{itemize}
    \item We propose SyncAnimation, a rendering framework for audio-driven expression and upper body generation. This framework generates an avatar that is highly consistent with the audio and displays a diversity of actions, while supporting both one-shot and zero-shot inference.
    
    \item We introduce the Audio2Pose and Audio2Emotion modules to support end-to-end efficient training, enabling high-precision poses and expression generation, and progressively generating audio-sync upper body, head, and lip shapes.
    
    \item Our algorithm achieves 41 FPS inference on an NVIDIA RTX 4090 GPU, and to our knowledge, this is the first real-time audio-driven avatar method capable of generating audio-sync upper body movement and head motion.

    \item Extensive experimental results show that SyncAnimation successfully generates realistic avatars with the same scale as the original video and significantly outperforms existing state-of-the-art methods in both quantitative and qualitative evaluations.
\end{itemize}

\section{Related Work}

\subsection{Person-Specific Paste-Back Generation}

GAN-based and NeRF-based methods are representative of paste-back generation. Among them, GAN-based talking head synthesis has primarily focused on generating video streams for the lip region, creating new visual effects for talking head avatar\cite{cheng2022videoretalking,zhang2023dinet,zhong2023identity,tan2025edtalk}. For example, Wav2Lip~\cite{prajwal2020lip} introduced a powerful lip-sync discriminator to supervise lip movement and penalize mismatched mouth shapes. IP-LAP~\cite{zhong2023identity} proposed an audio-to-landmark generator and a landmark-to-video model, using prior landmark and appearance information to reconstruct lips from a reference image. Recently, EdTalk presented an effective disentanglement framework, using orthogonal bases stored in a dedicated library to represent each spatial component for efficient audio-driven synthesis. With the rise of NeRF, earlier works~\cite{guo2021ad,ye2023geneface,li2023efficient,peng2024synctalk} have integrated NeRF into the task of synthesizing talking heads, using audio as the driving signal. For instance, AD-NeRF~\cite{guo2021ad} was the first to render both the torso and head of a person but suffered from poor generalization, and the synthesized lip movement sometimes appeared unnatural. ER-NeRF~\cite{li2023efficient} innovatively introduced tri-plane hash encoders and a region attention module, advocating a fast and precise lip-sync rendering approach. Geneface~\cite{ye2023geneface} and SyncTalk~\cite{peng2024synctalk} generated a generalized representation based on extensive 2D audiovisual datasets, ensuring synchronized lip movement across different audios.

The GAN-based methods map audio to lip sync, but other parts are pasted back. As shown in Fig.\ref{fig:Qualitative}, the reconstructed lip region appears blurry. In contrast, NeRF-based methods perform full-face synthesis but fail to synchronize audio with facial expressions, head poses, and upper-body movement.

\subsection{Pre-trained, Multi-person Joint Generation}
Leveraging the fundamental principles of text-to-image diffusion models, recent advances in video generation based on diffusion techniques have shown promising results~\cite{xu2024hallo,wang2024v,chen2024echomimic}. V-Express~\cite{wang2024v} effectively links lip movement, facial expressions, and head poses through progressive training and conditional dropout operations, enabling precise control using audio. Hallo~\cite{xu2024hallo} adopts a hierarchical audio-driven visual synthesis approach, achieving lip synchronization, expression diversity, and pose variation control. EchoMimic~\cite{chen2024echomimic} employs a novel training strategy that incorporates both audios and facial landmarks for avatar synthesis.

SD-based methods generate audio-driven talking avatars with poses and facial animations. However, large-scale training often results in poor resemblance to the original individual and comes with high computational costs, making real-time applications like live streaming and video conferencing challenging. For instance, generating a one-minute video can take up to half an hour. Additionally, these methods struggle with audio-motion mismatches when handling out-of-domain audio, further limiting their suitability for real-time use.

\begin{figure*}[h]
    \begin{minipage}{\linewidth}
		\centering
		\includegraphics[width=0.99\linewidth]{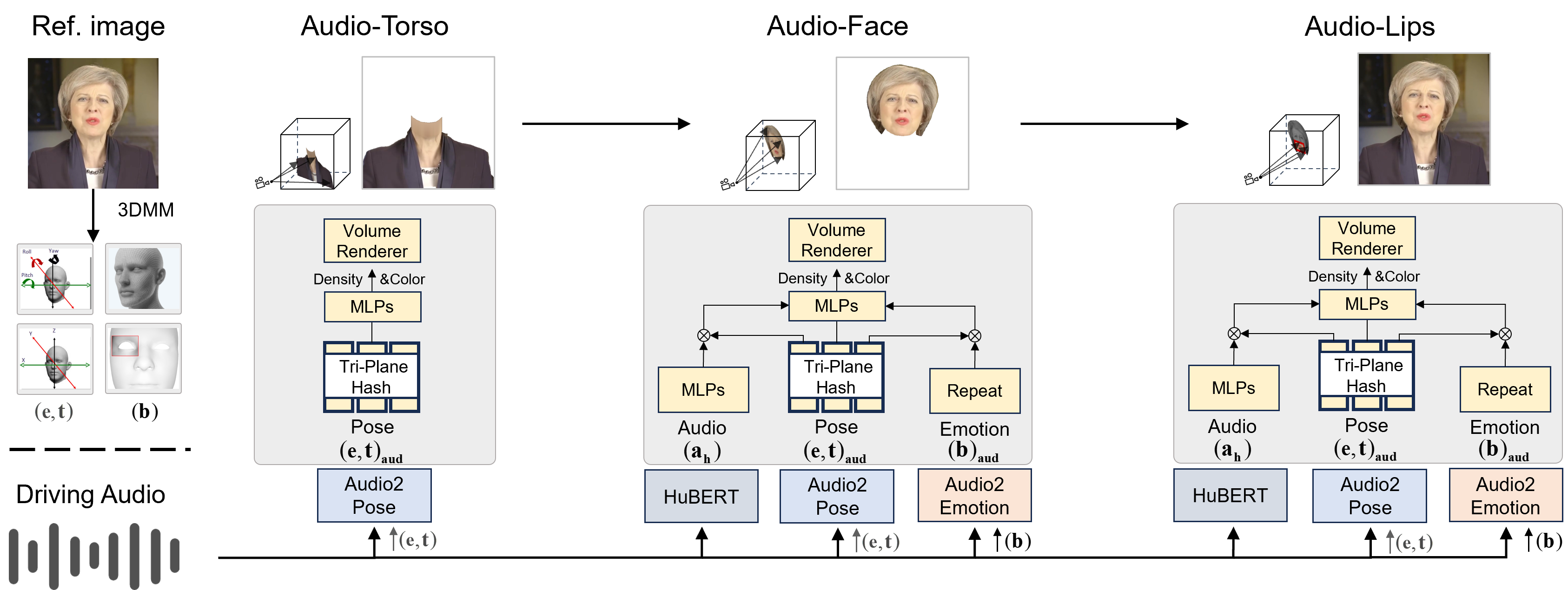}
    \end{minipage}
    \caption{Illustration of SyncAnimation Framework: Given a single image and audio input, the preprocessing stage extracts 3DMM parameters in NeRF space as references for Audio2Pose and Audio2Emotion (or alternatively, noise). The framework then progressively generates the upper body, head, and final lip refinement. Audio2Pose ensures poses consistency with the audio for upper-body generation, while Audio2Emotion aligns facial expression with the audio for head rendering, in conjunction with the generated poses.}
    \label{fig:pipeline}
\end{figure*}

\section{Method}

As shown in Fig.~\ref{fig:pipeline}, our method, SyncAnimation, is a jointly generative, audio-driven model that creates avatar including both upper-body and head. It emphasizes the correlation between audio and head poses, the consistency between head motion and upper-body movement, and the synchronization of audio with facial expressions and lip movement. The framework consists of three key components: (1).the AudioPose Syncer, which ensures the stable and accurate mapping of audio to dynamic head poses,in Sec.~\ref{sec:audioPose}. (2).the AudioEmotion Syncer, facilitating highly controllable and adaptable facial expressions driven by audio, in Sec.~\ref{sec:audioEmotion}. and (3).the High-Synchronization Human Renderer, responsible for rendering synchronized frames and ensuring seamless, realistic upper-body generation without the need for head pasting, as described in Sec.~\ref{sec:lipsync}. The following subsections explore the details of these three key modules.

\begin{figure}[ht]
    \begin{minipage}{\linewidth}
		\centering
		\includegraphics[width=0.95\linewidth]{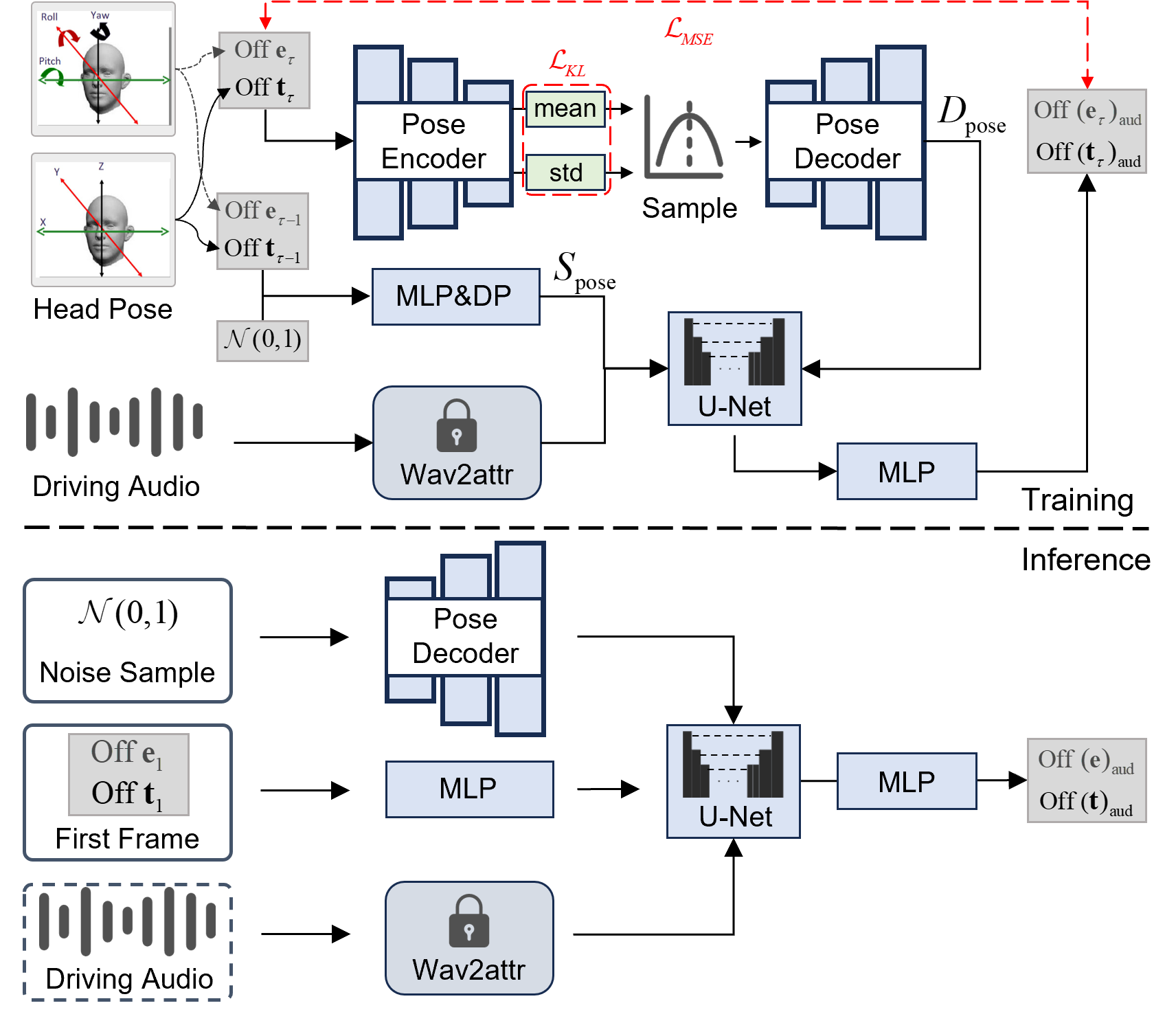}
    \end{minipage}
    \caption{Audio2Pose is designed to  reconstruct stable head pose offsets (\(\mathbf{Off}\text{ }(\mathbf{e}\)) and \(\mathbf{Off}\text{ }(\mathbf{t})\)) using audio and monocular input. Where, Wav2attr, a pre-trained audio encoder, is employed to encode audio vectors containing character-specific information. Additionally, a Gaussian-based VAE is integrated to introduce a diversity template \(\mathbf{S}_{\text{pose}}\), while a stability model \(\mathbf{D}_{\text{pose}}\) is implemented based on the poses labels with high dropout rate (\(\text{DP}=0.6\)), improving the effectiveness of pose reconstruction.}
    \label{fig:audio2Pose}
\end{figure}

\subsection{AudioPose Syncer}
\label{sec:audioPose}

Existing NeRF-based models, such as AD-Nerf, GenFace, and Synctalk, commonly rely on 3DMM estimators to predict the poses of a person to generate realistic avatar with lip-syncing. However, they overlook the intrinsic relationship between poses and audio, merely pasting the generated head back onto the original torso. This paste-back generation method results in head pose variations identical to the original video, independent of the audio.

\paragraph{Audio-aware Poses Generation.}
\label{PoseGeneration}
Learning a model that generates precise control over head motion from audio is extremely challenging for two issues: (1). From a theoretical standpoint, the rotation matrix controlling head motion must be orthogonal and have a determinant of 1. Direct prediction by the model may inevitably output non-orthogonal matrices during training, leading to errors in rotation calculations and affecting the final pose reconstruction; (2). From a practical perspective, there is an inherent ambiguity between audio and motion, rather than a one-to-one relationship. Phenomenologically, this is manifested in generated avatar that exhibit jitter or even frame skipping.

The proposed AudioPose aims to address these issues, as shown in Fig.\ref{fig:audio2Pose}. For issue 1, considering the difficulty in predicting orthogonal rotation matrices with a high-dimensional compact representation, we transform the task into predicting three scalar Euler angles (roll \(\alpha\), pitch \(\beta\), and yaw \(\gamma\)). Given rotation matrix \(\mathbf{R}\), the conversion formula is:
\begin{align}
    \beta &= \text{atan2}(\mathbf{R}_{32}, \mathbf{R}_{33}), \\
    \gamma &= \text{atan2}(-\mathbf{R}_{31}, \sqrt{\mathbf{R}_{32}^2 + \mathbf{R}_{33}^2}), \\
    \alpha &= \text{atan2}(\mathbf{R}_{21}, \mathbf{R}_{11})
\end{align}
\( \mathbf{R}_{\text{ij}} \) denotes the element in the \( i \)-th row and \( j \)-th column of the rotation matrix \( \mathbf{R} \). The Euler angles \(\mathbf{e} = (\alpha, \beta, \gamma)\) and the translation \(\mathbf{t} = (x, y, z)\) jointly define the head poses, controlling its orientation and position.

The non-linear transformation of the rotation matrix introduces new challenges, such as gimbal lock (\(\beta\) \(\approx \) \(\pm 90^\circ\)) causing non-invertibility, and pose variations influenced by Euler angles and translations within narrow ranges. In real-world videos, head poses are confined to small ranges, not extending to extreme values like \(\pm 90^\circ\) or \(\pm 180^\circ\). This insight is incorporated into training to address these challenges. Specially, the prediction range is restricted from \((\mathbf{e}, \mathbf{t}) \in (-\infty, +\infty)\) to \((\Delta \mathbf{e}, \Delta \mathbf{t}) \in \left(\bar{\mathbf{e}} \pm \Delta, \bar{\mathbf{t}} \pm \Delta \right)\), where \(\bar{\mathbf{e}}\) and \(\bar{\mathbf{t}}\) are the average poses, and \(\Delta\) defines the deviation. To ensure consistency, outputs are converted into a normalized distribution \((\bar{\Delta \mathbf{e}}, \sigma_{\Delta \mathbf{e}})\), standardizing the deviation range as follows: 
\begin{equation} 
    \textbf{Off}\text{ }\mathbf{(e)} = \frac{\Delta \mathbf{e} - \bar{\Delta \mathbf{e}}}{\sigma_{\Delta \mathbf{e}}}, \quad \textbf{Off}\text{ }\mathbf{(t)} = \frac{\Delta \mathbf{t} - \bar{\Delta \mathbf{t}}}{\sigma_{\Delta \mathbf{t}}}
\end{equation}
Next, we introduce two conditional vectors, \(\mathbf{S}_{\text{pose}}\) and \(\mathbf{D}_{\text{pose}}\), to address the ambiguity between audio and poses in issue 2. The pipeline for the poses prediction of Audio2Pose is:

\begin{equation} 
    \textbf{Off}\text{ }\mathbf{(e)}_{\text{aud}}, \textbf{Off}\text{ }\mathbf{(t)}_{\text{aud}} = F(g(\mathbf{a}), \mathbf{D}_{\text{pose}}, \mathbf{S}_{\text{pose}})
\end{equation}
Where \(F(\cdot)\) represents the poses generation model. Both \(\mathbf{D}_{\text{pose}}\) and \(\mathbf{S}_{\text{pose}}\) serve as conditional vectors, with \(\mathbf{S}_{\text{pose}}\) enhancing poses stability and \(\mathbf{D}_{\text{pose}}\) guiding its diversity. For \(g(\mathbf{a})\), we use FaceXHuBERT~\cite{FaceXHubert} as the audio encoder \(g(\cdot)\), as it is better suited for our 3D facial animation task compared to widely used pre-trained models like DeepSpeech and wav2vec~\cite{hannun2014deep,wav2vec}, which lack person-specific information.

Audio-only poses prediction \(\big(\mathbf{a} \rightarrow \textbf{Off}\text{ }\mathbf{(e)}_{\text{aud}}, \textbf{Off}\text{ }\mathbf{(t)}_{\text{aud}}\big)\) often lacks sufficient information to resolve the audio-to-pose ambiguity. To address this, we introduce an additional conditional vector \(\mathbf{S}_{\text{pose}}\) to enhance pose stability by providing a reference poses, reducing the feasible solution space from one-to-many to many-to-one. During training, the regression loss is backpropagated to update \(\mathbf{S}_{\text{pose}}\), which is adaptively adjusted to resolve ambiguities in the mapping. Specifically, at the \(\tau\)-th frame, the input is formed by adding Gaussian noise to the poses of the \((\tau-1)\)-th frame, which is then encoded through several Multilayer Perceptrons (MLPs) to generate \(\mathbf{S}_{\text{pose}}\). The process is expressed as follows.
\begin{equation}
    \mathbf{S}_{\text{pose}} = f_{\text{MLPs}}\big([\mathbf{\textbf{Off}\text{ }\mathbf{(e_{\tau-1})}, \textbf{Off}\text{ }\mathbf{(t_{\tau-1})}}] + \mathcal{N}(\mu_S, \delta_S) \big)
\end{equation}
where $\textbf{Off}\text{ }\mathbf{(e_{\tau-1})}$ and $ \mathbf{\textbf{Off}\text{ }\mathbf{(t_{\tau-1})}}$ represent the pose information of the $(\tau-1)$-th frame, and $\mathcal{N}(\mu_S, \delta_S)$ denotes Gaussian noise with mean $\mu_S$ and standard deviation $\delta_S$.

Using \(\mathbf{S}_{\text{pose}}\), stable poses are generated. To prevent shortcut learning~\cite{geirhos2020shortcut}, where the model directly maps inputs to outputs, we add high dropout during \(\mathbf{S}_{\text{pose}}\) generation and introduce a diversity-guided conditional vector \(\mathbf{D}_{\text{pose}}\) via a Variational Autoencoder (VAE). \(\mathbf{D}_{\text{pose}}\) introduces uncertainty, preventing the model from outputting static poses, and provides diverse pose templates through Gaussian sampling. During training, VAE backpropagation ensures faster convergence and improved accuracy compared to single audio regression, as shown in Exp. \ref{sec:pose_effectiveness}.

To normalize the diversity-guided \(\mathbf{D}_{\text{pose}}\) vector space, we apply the KL-divergence loss:
\begin{equation}
    \mathcal{L}_\text{KL}=D_{\text{KL}}\big(\mathcal{N}(\boldsymbol{\mu_D}, \boldsymbol{\delta^2_D})||\mathcal{N}(0,\mathbf{I})\big)
\end{equation}
where $\mathcal{N}(\boldsymbol{\mu_D}, \boldsymbol{\delta^2_D})  \in \mathcal{R}^D$ is the latent distribution predicted by the VAE encoder, and \(\mathcal{N}(0, \mathbf{I})\) represents the standard normal distribution used as a prior. This loss ensures that the \(\mathbf{D}_{\text{pose}}\) vector space remains well-behaved and regularized, facilitating stable and diverse poses generation.

For the poses generation model \(F(\cdot)\), we use the U-Net architecture \cite{qian2021speech,wang2024v,chen2024echomimic,xu2024hallo}, which leverages multi-scale feature extraction to incorporate audio features, the diversity-guided conditional vector \(\mathbf{D}_{\text{pose}}\), and the stability constraint vector \(\mathbf{S}_{\text{pose}}\). The U-Net produces a shared feature representation, which is then processed by MLPs to predict rotation \(\textbf{Off}\text{ }\mathbf{(e)}_{\text{aud}}\) and translation \(\textbf{Off}\text{ }\mathbf{(t)}_{\text{aud}}\). A reconstruction loss \(\mathcal{L}_\text{reg}\) is applied to align the generated poses with the ground truth.
\begin{equation}
    \mathcal{L}_\text{reg}=\Vert \big(\textbf{Off}\text{ }\mathbf{(e)}, \textbf{Off}\text{ }\mathbf{(t)}\big) - \big(\textbf{Off}\text{ }\mathbf{(e)}_{\text{aud}}, \textbf{Off}\text{ }\mathbf{(t)}_{\text{aud}}\big) \Vert_1
\end{equation}
Therefore, the final poses generation loss is:
\begin{equation}
   \mathcal{L}_\text{pose}=\lambda_{\text{KL}}\mathcal{L}_\text{KL}+\lambda_{\text{reg}}\mathcal{L}_\text{reg}.
\end{equation}
where $\lambda_{\text{KL}}$ and $\lambda_{\text{reg}}$ are weights applied to the VAE and regression loss terms. We set $\lambda_{\text{KL}} = 0.1$ and $\lambda_{\text{reg}}=1$ in our experiments.

\paragraph{Audio-guided Upper-body Generation.}
\label{Torso Generation}
We generally follow the rendering process in preceding work~\cite{guo2021ad,li2023efficient}. Originated from the representation of neural radiance field, the implicit function is defined as $\mathcal{F}_\text{NeRF}:(\text{x}, d)\rightarrow(c, \sigma)$, where $\text{x}=(x,y,z)$ refers to 3D spatial location and $d$ is the viewing direction. The output $c$ and $\sigma$ determine color value and corresponding density, based on which we can attain the pixel color values $\hat{\mathcal{C}}$ accumulated through rays $r(t)=o+td$ dispatched from camera center $o$ via:
\begin{equation}
    \hat{\mathcal{C}}(r)=\int_{t_n}^{t_f}\sigma(r(t)\cdot c(r(t),d)\cdot T(t)dt,
\end{equation}
where $t_n$ and $t_f$ represent the near and far bounds, with $T(t)$ measuring cumulative transmittance from $t_n$ to $t$ formulated as:
\begin{equation}
    T(t)=exp\left(-\int_{t_n}^t\sigma(r(s))ds\right).
\end{equation}
To render more high-fidelity and realistic scene efficiently, we employ a 2D-multiresolution hash encoder introduced by~\cite{hashencoding,li2023efficient}. A specific encoder designed for position $\text{x}=(x,y,z)$ projected on plane $AB$ is defined as $\mathcal{H}^{AB}:(a,b)\rightarrow h^\text{AB}_{ab}$, where $a$ and $b$ are projected coordinates, while $h^\text{AB}_{ab}\in \mathbb{R}^{LD}$ represents $L$ levels of features with $D$ dimensions. 

Since we pursue dynamic upper-body generation and consistency between head motion and upper body movement, we recover the audio-predicted $\mathbf{\textbf{Off}\text{ }\mathbf{(e)}}_{\text{aud}}$  and $\mathbf{\textbf{Off}\text{ }\mathbf{(t)}}_{\text{aud}}$ to original space representations, $\mathbf{e}_\text{aud}$ and $\mathbf{t}_\text{aud}$, using inverse normalization. Subsequently, three trainable coordinates $\Tilde{X}$ are deformed as rendering conditions. Consequently, the implicit function is given by:
\begin{equation}
    F_\text{upper-body}:(\text{x},\mathbf{e}_\text{aud},\mathbf{t}_\text{aud},\tilde{X},\mathcal{H}^{t})\rightarrow(c,\sigma),
\end{equation}
where $\mathcal{H}^t$ denote the hash encoder. During training, we optimize the upper-body model by minimizing the error between $\hat{\mathcal{C}}(r)$ and genuine pixel color $\mathcal{C}(r)$ through:
\begin{equation}
    \mathcal{L}_\text{upper-body}=||\mathcal{C}(r)-\hat{\mathcal{C}}(r)||^2_2.
\end{equation}
\begin{figure}[ht]
    \begin{minipage}{\linewidth}
		\centering
		\includegraphics[width=0.95\linewidth]{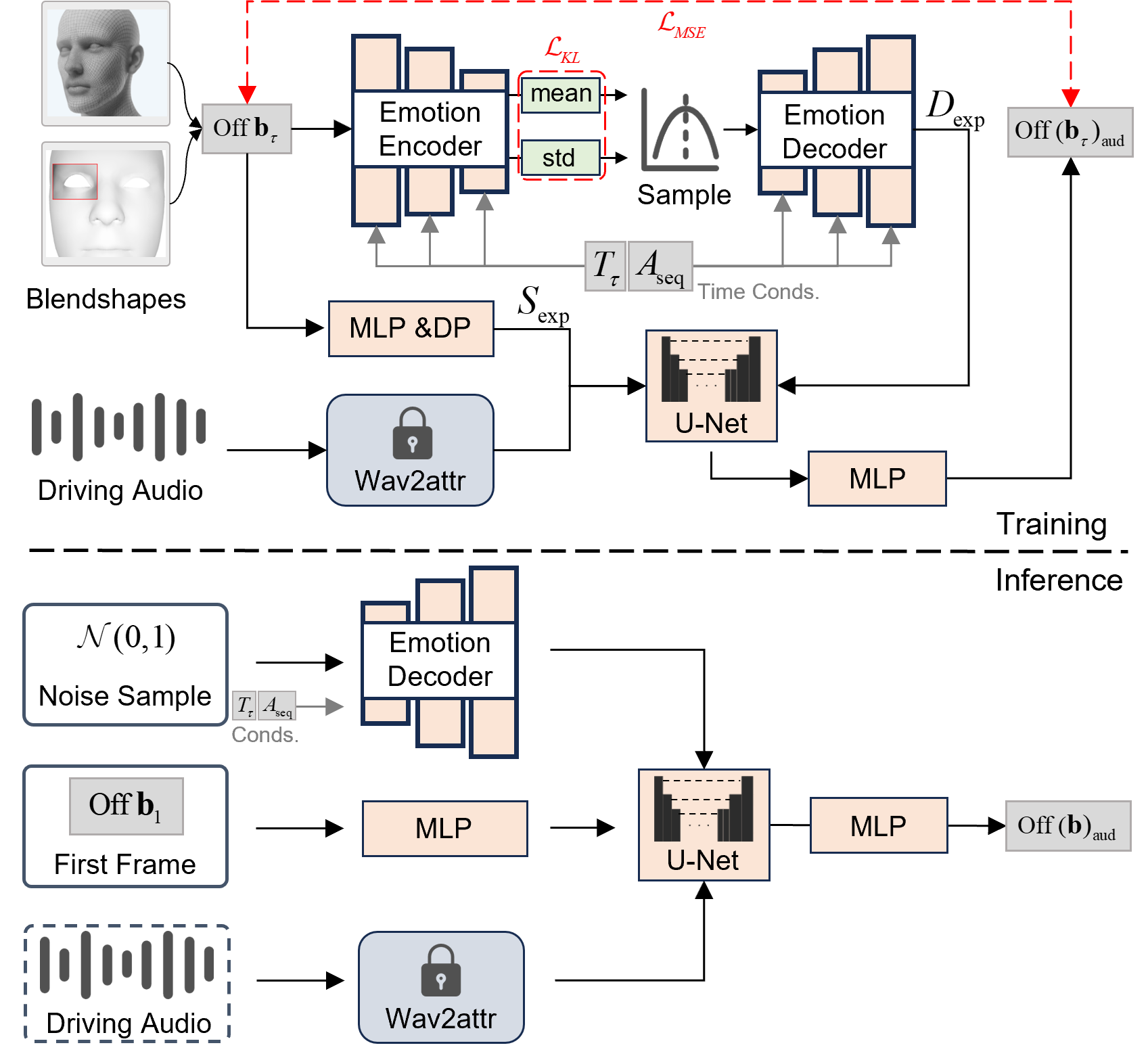}
    \end{minipage}
    \caption{Audio2Emotion is designed to learn and reconstruct 3DMM expression offsets (\(\mathbf{Off}\text{ }(\mathbf{b})\)) using audio and monocular input. The structure is similar to that of Audio2Pose, but due to the weak correlation between facial expressions and audio, and the periodic nature of blinking, we modify the diversity template \(\mathbf{S}_{\text{exp}}\). It is replaced by a conditional VAE guided by periodic time features \(\mathbf{T}_{\tau}\) and context-dependent audio features \(\mathbf{A}_{\text{seq}}\).}
    \label{fig:audioEmotion}
\end{figure}
\subsection{AudioEmotion Syncer}
\label{sec:audioEmotion}
\paragraph{Lifelike Expression Prediction.}
Audio-driven research primarily focuses on lip-sync, neglecting other facial expressions with weaker audio correlation~\cite{tian2025emo}, leading to unnatural expressions. We first use arkit face blendshapes, semantically rich 3D coefficients, to model the upper face region \(\mathbf{b}\) \cite{peng2023emotalk}.

However, For the eyes on the upper face, compared to the critical cases of head poses, the values of \(\mathbf{b}\) approach extreme values (0 or 1) during fully open and fully closed eyes. To enable full eye closure, we predict the deviation \(\textbf{Off}\text{ }\mathbf{(b)}\) relative to the average \(\bar{\mathbf{b}}\), and incorporate stability constraints \(\mathbf{S}_{\text{exp}}\) and diversity guidance \(\mathbf{D}_{\text{exp}}\), as shown in Fig.\ref{fig:audioEmotion}. The formulation is expressed as follows:
\begin{equation} 
    \textbf{Off}\text{ }\mathbf{(b)}_{\text{aud}} = F(g(\mathbf{a}), \mathbf{D}_{\text{exp}}, \mathbf{S}_{\text{exp}})
\end{equation}
Blinking strongly relies on temporal flow and exhibits periodicity, which results in slow convergence and low fitting accuracy, causing unnatural behavior, such as characters failing to blink. To address this, we modified the VAE within the diversity-guided conditional vector \(\mathbf{D}_{\text{exp}}\) to capture temporal dependencies and strengthen the correlation with audio. By incorporating time features \(\mathbf{T}_{\tau}\) and time-dependent audio features \(\mathbf{A}_{\text{seq}}\), the CVAE generates \(\mathbf{D}_{\text{exp}}\) with strong temporal correlations, as formulated below:
\begin{equation}
    \mathbf{D}_{\text{exp}} = f_{\text{CVAE}}\big(\textbf{Off}(\mathbf{b}) \mid T_{\tau}, \mathbf{A}_{\text{seq}}\big)
\end{equation}
where \(\mathbf{T}_{\tau}\) is encoded using sinusoidal periodic encoding, processed through a MLP. \(\mathbf{A}_{\text{seq}}\) is derived by integrating audio data from \(n\) neighboring frames \(\{\mathbf{a}_{\tau-n}, \dots, \mathbf{a}_{\tau+n}\}\), and then processed through stacked convolutional layers to generate temporally contextualized representations that ensure dynamic expression continuity.

Meanwhile, considering the weak correlation between audio and expressions, we provide a more accurate expression reference template by modifying the input for predicting the stability constraint vector \(\mathbf{S}_{\text{exp}}\). Specifically, we replace the previous frame's expression information \(\mathbf{b}_{\tau-1}\) with added Gaussian noise \(\mathcal{N}(\mu_S, \delta_S)\) for the current frame's expression \(\mathbf{b}_{\tau}\). The formulation is as follows:
\begin{equation}
    \mathbf{S} = f_{\text{MLPs}}\big(\mathbf{\textbf{Off}\text{ }\mathbf{(b_{\tau})}}\big)
\end{equation}
For the audio-driven expression model, we optimize \(L_\text{exp}\) using the following loss function:
\begin{equation}
    \mathcal{L}_\text{exp}=\lambda_{\text{KL}}\mathcal{L}_\text{KL}+\lambda_{\text{reg}}\mathcal{L}_\text{reg}.
\end{equation}
where $\lambda_{\text{KL}}$ and $\lambda_{\text{reg}}$ are weights applied to the CAVE and regression loss terms. We set $\lambda_{\text{KL}} = 0.1$ and $\lambda_{\text{reg}}=1$ in our experiments.

\paragraph{Dynamic Head Rendering.}

We use a technique similar to Sec.\ref{Torso Generation} to generate realistic head. The audio-predicted \(\textbf{Off}\text{ 
 }\mathbf{e}_{\text{aud}}\), \(\textbf{Off}\text{ }\mathbf{t}_{\text{aud}},\) and \(\textbf{Off}\text{  
 }\mathbf{b}_{\text{aud}}\) are converted to \(\mathbf{e}_{\text{aud}}, \mathbf{t}_{\text{aud}},\) and \(\mathbf{b}_{\text{aud}}\) in the original space via inverse normalization. Owing to the complexity of head generation, we utilize three 2D hash encoders, while each operates on a specific plane. By concatenating outcomes from three planes, we get the final geometric feature. We also use predicted coefficients to control expression. Moreover, we introduce audio feature to enhance head motion prediction. Subsequently, the implicit function can be defined as:

\begin{equation}
    F_\text{head}:(\text{x}, \mathbf{e}_\text{aud}, \mathbf{t}_\text{aud}, \mathbf{b}_\text{aud}, a_\text{h}, \mathcal{H}^3)\rightarrow(c,\sigma),
\end{equation}

where $a_\text{h}$ is audio representation extracted by Hubert ~\cite{Hubert}, which is different from what we use to predict poses and expression due to diverse task demands. The corresponding loss $\mathcal{L}_\text{head}$ is consistent with $\mathcal{L}_\text{upper-body}$. On account of our high-quality poses evaluation, head can match upper-body movement accurately and smoothly. 

\subsection{High-Synchronization Human Renderer}
\label{sec:lipsync}

\paragraph{Facial-aware attention.}
To better leverage audio and expression information in different facial region, we exploit channel-wise attention mechanism~\cite{li2023efficient}. Given the output of the hash encoders, we obtain audio and expression related attention weights through two MLP $\text{Attn}_\text{aud}$ and $\text{Attn}_\text{exp}$:
\begin{equation}
\begin{split}
    v_\text{aud}=\text{Attn}_\text{aud}(\mathcal{H}^3(x)) \\
    v_\text{exp}=\text{Attn}_\text{exp}(\mathcal{H}^3(x)).
\end{split}
\end{equation}
By applying hadamard product we attain region-aware features $a_{h,x}=a_{h,x}\odot v_\text{aud}$ and $\mathbf{b}_{\text{out},x}=\mathbf{b}_\text{out}\odot v_\text{exp}$. Such operation makes sure model can explore useful information in a disentangled way and thus increase rendering quality.

\paragraph{Fine Lip Optimization.}
Though audio feature benefits holistic head rendering, we seek to manually augment attention weight on lip region due to more relevance. By utilizing mask technique~\cite{peng2024synctalk}, we lower the attention weight out of lip area. Meanwhile, we use LPIPS loss focused on lip zone to gain finer result.

The process is expressed as:
\begin{equation}
    \mathcal{L}_\text{lip}=\mathcal{L}_\text{head}+ \lambda \text{LPIPS}(\mathcal{P},\hat{\mathcal{P}}).
\end{equation}
Consequently, we further optimize LPIPS loss in the lip region 


\subsection{Training Details}
\label{sec:Training}
We apply a three-stage training strategy, progressively optimizing audio-driven upper-body generation, realistic and expressive facial generation consistent with the upper body, and lip refinement. This approach ensures the creation of a jointly generative, audio-driven avatar, where each stage contributes to enhancing the audio consistency, realism, and coherence of the final generated avatar.

\textbf{Stage 1.} In stage 1, 
the focus is on optimizing audio-driven poses (Audio2Pose) and upper-body generation \(F_\text{upper-body}\) using the losses \(\mathcal{L}_\text{pose}\) and \(\mathcal{L}_\text{upper-body}\). 

\textbf{Stage 2.} In stage 2, 
the audio-driven expression model (Audio2Emotion) is optimized with \(\mathcal{L}_\text{exp}\), while incorporating the Audio2Pose model to jointly optimize head rendering generation \(F_\text{head}\). 

\textbf{Stage 3.} In stage 3 , 
with well-trained audio-driven poses and expression models, the lip region refinement is performed by extracting random patches \(\mathcal{P}\) from the head and integrating a homogenous LPIPS loss weighted by \(\lambda\) with \(\mathcal{L}_\text{head}\). 

Based on our well-designed training process, our generated avatars becomes stable and realistic.

\begin{table*}[]
\begingroup
\renewcommand{\arraystretch}{1}
\centering
\scalebox{1}{
\begin{tabular}{ll|llll|lll|ll}
\toprule
\multicolumn{2}{c|}{\multirow{2}{*}{\textbf{Method}}} & \multicolumn{4}{c|}{\textbf{Image Quality}} & \multicolumn{3}{c|}{\textbf{Lip Sync}} & \multicolumn{2}{c}{\textbf{Head Motion}} \\
\cline{3-11}
 & & PSNR$\uparrow$ & LPIPS$\downarrow$ & SSIM$\uparrow$ & FID$\downarrow$ & LMD$\downarrow$ & AUE$\downarrow$ & SyncScore$\uparrow$ & Diversity$\uparrow$ & EAR$\downarrow$ \\
\hline 
\multirow{4}{*}{GAN} & Wav2Lip & 18.8071 & 0.2523 & 0.6571 & 40.3604 & 5.2161 & 4.5499 & \textbf{9.2332} & 0.0782 & 0.0535 \\
& DINet  & 18.6620 & 0.2547 & 0.6515 & 34.0215 & 5.2784 & 4.4457 & 7.3138 & 0.0560 & 0.0568     \\ 
& IP-LAP  & 18.7763 & 0.2438 & 0.6531 & 34.8322 & 5.4905 & 4.7474 & 3.4331 & 0.0444 & 0.0545   \\ 
& EDTalk & 18.7482 & 0.2896 & 0.6670 & 53.7573 & 5.0925 & 4.8511 & 7.3092 & 0.1046 & 0.0441     \\ 
\hline 
\multirow{3}{*}{Nerf} & ER-NeRF & 18.8610 & 0.2323 & 0.6799 & 37.5604 & 3.7192 & 4.8899 & 6.1909 & 0.0935 & 0.0439    \\
& SyncTalk & 18.8035 & 0.2287 & 0.6766 & 32.9779 & 3.6106 & 3.6198 & 7.0865 & 0.0822 & 0.0512     \\ 
& GeneFace$++$ & 19.0344 & 0.2443 & 0.6819 & 40.5045 & 5.3823 & 4.0530 & 7.1118 & 0.0726 & 0.0504    \\ 
\hline
\multirow{3}{*}{SD} & Hallo & 17.9117 & 0.2678 & 0.6305 & 35.8346 & 6.0491 & 4.3952 & 7.2458 & 0.2166 & 0.0471     \\
& V-Express & 17.3918 & 0.2842 & 0.6121 & 46.6975 & 6.9958 & 5.4740 & 7.4739 & 0.1039 & 0.0490   \\ 
& EchoMimic & 14.4399 & 0.4524 & 0.4562 & 59.3261 & 8.0643 & 5.1434 & 6.0574 & 0.1864 & 0.0468     \\ \hline 
\multicolumn{2}{c|}{SyncAnimation-One}  & 21.2323 & 0.1543 & 0.7343 & \textbf{20.0567} & 3.2215 & \textbf{3.5485} & 7.1389 & 0.2570 & \textbf{0.0357} \\
\multicolumn{2}{c|}{SyncAnimation-Zero}  & \textbf{21.4006} & \textbf{0.1532} & \textbf{0.7411} & 21.7353 & \textbf{3.1878} & 3.5515 & 7.1499 & \textbf{0.2652} & 0.0386 \\ 
\bottomrule
\end{tabular}
}
\caption{Quantitative comparisons with state-of-the-art methods. "SyncAnimation-One" refers to one-shot inference, while "SyncAnimation-Zero" indicates zero-shot inference using noise of the same dimension. We achieve state-of-the-art performance on most metrics.}

\label{tab1:baseline}
\endgroup
\end{table*}

\section{Experiments}

\subsection{Implementations}

\noindent\textbf{Dataset.} To ensure a fair comparison, the experimental dataset was obtained from publicly available video collections in ~\cite{guo2021ad,li2023efficient,peng2024synctalk} and HDTF ~\cite{zhang2021flow}. We collected well-edited video sequences featuring speakers of English, French, and Korean, with an average of 6665 frames per video in 25 FPS. 
Each raw video was standardized to 512×512, with the a center portrait.

\noindent\textbf{Quantitative Evaluation Metrics} 
We demonstrate the superiority of our method on multiple metrics that have been commonly used in previous studies. For evaluating image quality, we employ full-reference metrics, including Peak Signal-to-Noise Ratio (PSNR) \cite{hore2010image}, Learned Perceptual Image Patch Similarity (LPIPS) \cite{zhang2018unreasonable}, Structural Similarity Index Measure (SSIM) \cite{wang2004image}, and Frechet Inception Distance (FID) \cite{heusel2017gans}. In terms of lip and face synchronization, we utilize landmark distance (LMD) to measure the synchronicity of facial movement \cite{chen2018lip}, and Action Unit Error (AUE) to evaluate the accuracy of facial expressions \cite{baltruvsaitis2015cross}. Furthermore, we introduce Lip Sync Error Confidence (LSE-C), consistent with Wav2Lip, to evaluate the synchronization between lip movement and audio \cite{prajwal2020lip}.For the diversity of head motion generated, a standard deviation of the embeddings of head motion features is extracted from the generated frames using Hopenet \cite{ruiz2018fine}. For eye blink detection, The Eye Aspect Ratio (EAR), calculated based on the positions of facial landmarks around the eyes, is utilized to evaluate the naturalness and accuracy of generated eye movement (e.g., eye closure) \cite{soukupova2016eye}.

\noindent\textbf{Comparison Baselines.} We compare SyncAnimation with two SOTA methods for lip synchronization and motion generation. For lip synchronization, we include GAN-based methods like Wav2Lip~\cite{prajwal2020lip}, DInet~\cite{zhang2023dinet}, and IP-LAP~\cite{zhong2023identity}, as well as NeRF-based methods such as SyncTalk~\cite{peng2024synctalk}, AD-NeRF~\cite{guo2021ad}, ER-NeRF~\cite{li2023efficient}, and GeneFace++~\cite{ye2023geneface++}. For motion generation, we compare SyncAnimation, the first NeRF-based method to perform zero-shot audio-driven generation, with three large-scale stable diffusion models: V-Express~\cite{wang2024v}, Hallo~\cite{xu2024hallo}, and EchoMimic~\cite{chen2024echomimic}.

\noindent\textbf{Implementation Details.}  We train SyncAnimation with separate steps for upper-body generation, head rendering, and lip-audio synchronization, using 15k, 12k, and 4k steps, respectively. Each iteration samples $256^2$ rays and employs a 2D hash encoder with parameters \(L = 14\) and \(F = 1\). The AdamW optimizer is used, with learning rates set to 0.01 for the hash encoder and 0.001 for the Audio2Pose and Audio2Emotion modules. The total training time is approximately 4 hours on an NVIDIA RTX 4090 GPU.

\subsection{Quantitative Evaluation}
\paragraph{Compare with baseline.}

We first compare one-shot SyncAnimation (SyncAnimation-One) with several state-of-the-art methods, driving using the first frame and test audio for driving. As shown in Tab.\ref{tab1:baseline}, our method outperforms GAN-based, NeRF-based, and SD-based approaches in several aspects.(1) In terms of image quality, SyncAnimation-One achieves substantial improvements, with leading performance on PSNR, LMD, and FID metrics, indicating its ability to generate high-quality and detail-preserving avatars in audio-driven tasks. (2) For synchronization, SyncAnimation-One outperforms other methods on LMD and AUE metrics, demonstrating its capability to achieve excellent audio-lip synchronization. While SyncScore falls slightly behind GAN-based methods on the SyncScore metric due to their use of SyncNet as a training loss, SyncAnimation still surpasses NeRF-based methods. (3) Regarding the diversity of head motion and expression generated from audio-driven inputs, SyncAnimation-One significantly outperforms GAN-based and NeRF-based, and even SD-based methods in one-shot inference. It also achieves the smallest absolute error between predicted and ground truth EAR. This advantage arises from SyncAnimation-One's ability to dynamically render individuals based on audio cues, rather than solely relying on head mapping.

Our framework supports not only one-shot inference but also zero-shot inference. Zero-shot SyncAnimation(SyncAnimation-Zero) replaces the reference poses and blendshape with gaussian noise of the same dimensions and compare the results. The comparison is shown in Tab.\ref{tab1:baseline}, where SyncAnimation-One (using the first frame as reference) and SyncAnimation-Zero (using Gaussian noise as reference) achieve similar results across all metrics and consistently outperform other methods. This demonstrate that SyncAnimation does not rely on reference inputs for identity or information during inference. Instead, it operates as a highly audio-correlated, audio-driven framework, achieving dynamic rendering based on audio cues alone.

\subsection{Qualitative Evaluation}

\begin{figure*}[ht]
    \begin{minipage}{\linewidth}
		\centering
		\includegraphics[width=1\linewidth]{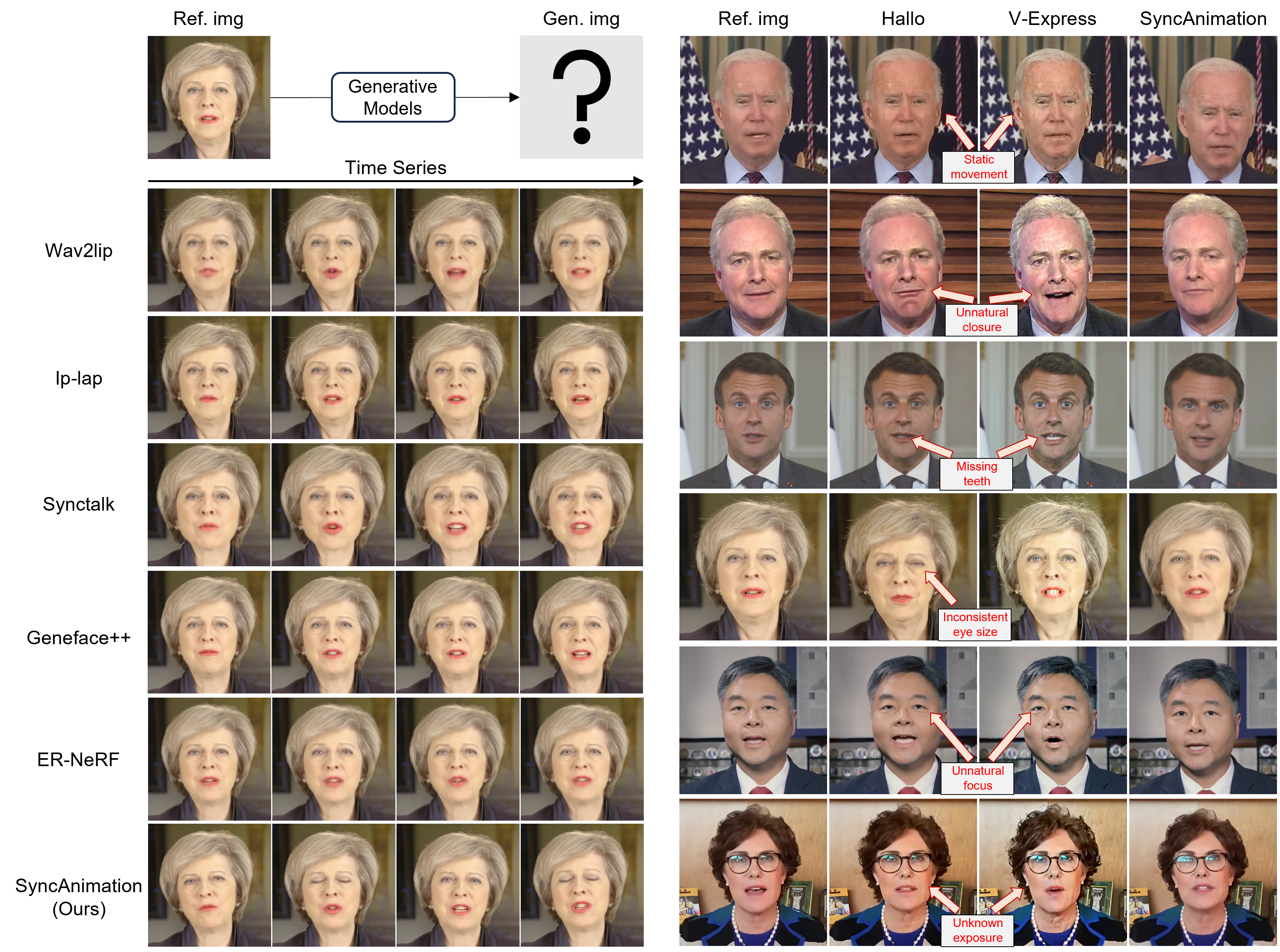}
    \end{minipage}
    \caption{Visual comparison with outputs of baselines. GAN-based and NeRF-based methods generate avatar with fixed poses and expressions, while SD-based methods introduce expression changes but lack face detail and pose movement. SyncAnimation uniquely achieves jointly generative, audio-driven realistic expressions and movable poses.}
    \label{fig:Qualitative}
\end{figure*}

In the previous section, we quantitatively demonstrated the superiority of SyncAnimation across multiple metrics. In this section, we visually assess the quality of generated frames by comparing GAN-based, NeRF-based, and SD-based methods with our SyncAnimation, as illustrated in Fig.\ref{fig:Qualitative}. Unlike other methods, which can render the upper body in single-frame audio-driven scenarios but fail to synchronize movement with audio, SyncAnimation excels by producing upper body with natural, audio-sync movement. This success is attributed to the AudioPose Syncer module and its joint generative design. Moreover, GAN-based and NeRF-based approaches struggle to manage upper facial expressions during one-shot inference. Although SD-based methods can generate diverse facial expressions, their reliance on large-scale training often prevents them from capturing fine-grained details like the eyes and lips, leading to issues such as asymmetrical eyes, overexposure, missing teeth, unfocused eyes, and unnatural lip closures. In contrast, SyncAnimation not only preserves the subject’s identity with superior fidelity and resolution but also accurately reproduces subtle actions, such as blinking and eyebrow movement, thanks to its AudioEmotion Syncer module. SyncAnimation achieves not only an overwhelming quantitative advantage over SD-based methods in inference but also outperforms them qualitatively in facial details and upper-body movement. We recommend watching the supplementary video for a more comprehensive comparison.

\begin{table*}[]
\begingroup
\renewcommand{\arraystretch}{1}
\centering
\scalebox{1}{
\begin{tabular}{l|llll|lll|ll}
\toprule
\multirow{2}{*}{\textbf{Method}} & \multicolumn{4}{c|}{\textbf{Image Quality}} & \multicolumn{3}{c|}{\textbf{Lip Sync}} & \multicolumn{2}{c}{\textbf{Head Motion}} \\
\cline{2-10}
 & PSNR & LPIPS & SSIM & FID & LMD & AUE & SyncScore & Diversity & EAR \\
\hline 

SyncAnimation  & 20.1038 & 0.1860 & 0.6752 & 26.8092 & 3.0746 & \textbf{2.8812} & \textbf{7.8717} & \textbf{0.2443} & \textbf{0.0405}  \\
SyncAnimation-MedScale   & 20.2972 & 0.1685 & 0.7051 & 27.4110 & 2.7120 & 3.7966 & 7.2024 & 0.2158 & 0.0413 \\
SyncAnimation-MaxScale &\textbf{21.5558} & \textbf{0.1192} & \textbf{0.7761} & \textbf{22.3477} & \textbf{2.0754} & 4.0455 & 7.3364 & 0.1632 & 0.0406 \\ 
\bottomrule
\end{tabular}
}
\caption{Qualitative comparison with varying upper-body scales}
\label{tab2:torsoScaling}
\endgroup
\end{table*}

\begin{figure}[ht]
    \begin{minipage}{\linewidth}
		\centering
		\includegraphics[width=1.\linewidth]{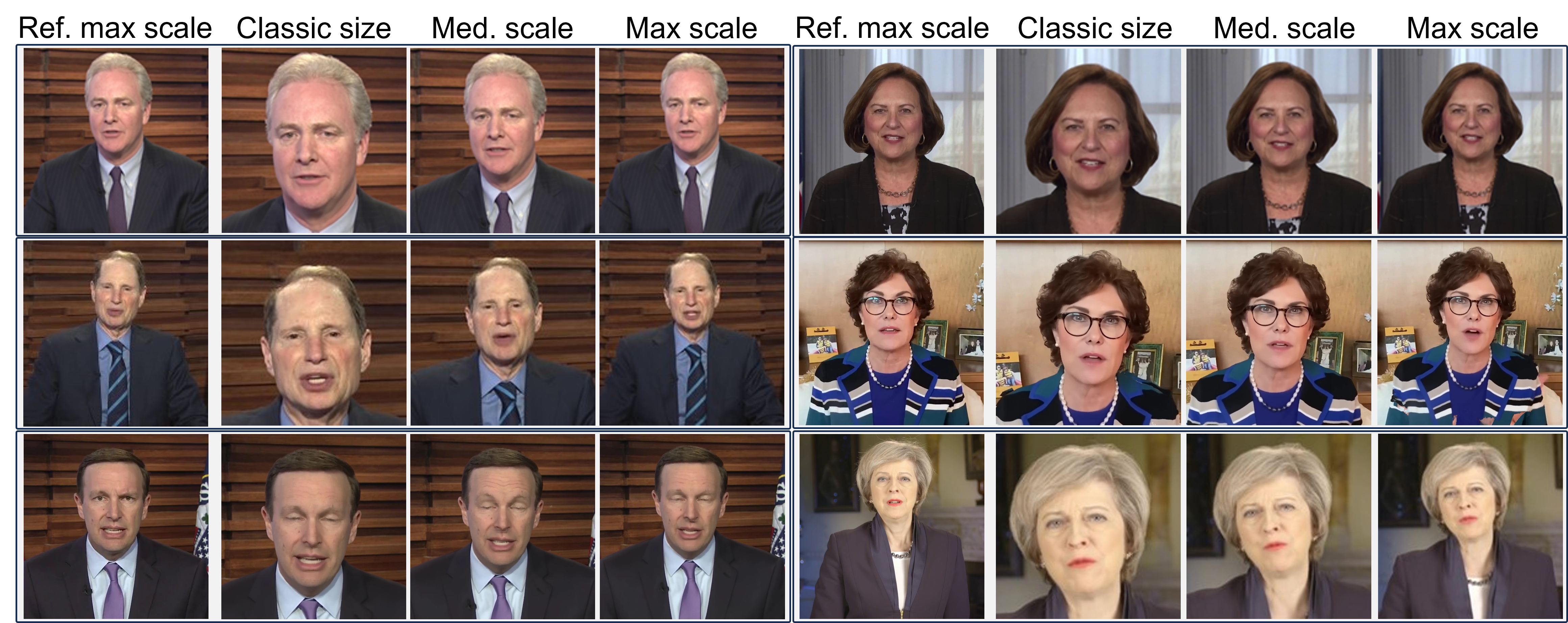}
    \end{minipage}
    \caption{Visual generation results of the proposed method given different upper-body scaling expansion}
    \label{fig:torso_scaling}
\end{figure}

\subsection{Upper-body Scaling Expansion}
In practical real-time applications (e.g., news broadcasting, live teaching), individuals often appear as upper body. Existing methods~\cite{li2023efficient,ye2023geneface,ye2023geneface++,peng2024synctalk} typically paste the rendered head back onto the torso of the original frame in the final step. This operation restricts the freedom of movement, making the output appear unnatural and limiting its applicability. Consequently, there is a growing demand for direct upper-body rendering that is highly synchronized with audio. 

In this chapter, we break away from the conventional approach of restricting upper-body scale and reject the practice of pasting back to the original frame. Instead, we directly render upper-body avatars driven by audio. Our method, SyncAnimation , progressively increases the proportion of the upper body in the rendered images. The metrics for image quality and audio consistency are shown in Tab.\ref{tab2:torsoScaling}. We observe that as the upper-body scale increases, the overall image quality improves, but there is a decline in lip-sync consistency and head motion diversity. Based on this, we conclude that the upper body is easier to render compared to complex face details, with a higher upper-body proportion resulting in better overall image quality. However, as the focus of rendering shifts more towards the upper body, the reduced attention on the head leads to a decrease in the other two metrics. The EAR remains relatively unchanged due to the periodic time features in \(\mathbf{S}_{\text{exp}}\) introduced by the proposed Audio2Emotion. Furthermore, as illustrated in Fig.\ref{fig:torso_scaling}, we scale the rendered upper-body avatar to match the size in the original video. The above demonstrates that SyncAnimation can directly generate upper-body avatars with strong audio correlation and natural, unrestricted poses.

\subsection{Ablation Study}
\begin{figure}[ht]
    \begin{minipage}{\linewidth}
		\centering
		\includegraphics[width=1\linewidth]{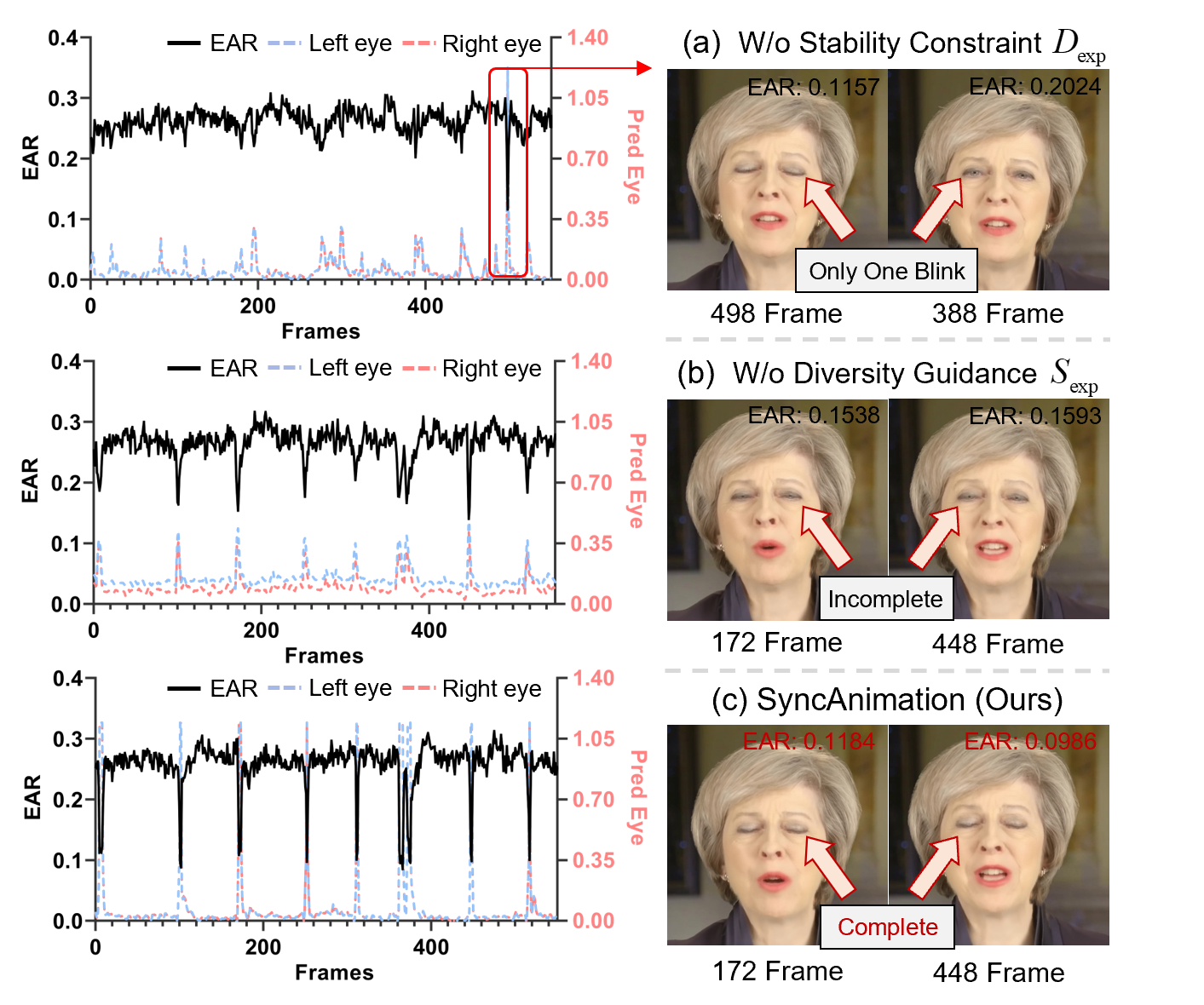}
    \end{minipage}
    \caption{Ablation Study on $\mathbf{D_{\text{exp}}}$ and $\mathbf{S_{\text{exp}}}$ in OOD audio inference. Removing them will result in (a) and (b).}
    \label{fig:ab_EXP}
\end{figure}

In this section, we report ablation studies under the joint generation setting to validate the effectiveness of our major contributions from two perspectives. Additionally, to demonstrate the adaptability of the SyncAnimation to Out-of-Domain (OOD) audio, we evaluate the impact of different backbones on rendering structures under external audio conditions. The results are presented in Tab.\ref{AudioPose ab} and Fig.\ref{fig:ab_EXP}.

\subsubsection{Effectiveness of $\mathbf{D_{\text{pose}}}$ and $\mathbf{S_{\text{pose}}}$}
\label{sec:pose_effectiveness}
The Audio2Pose model plays a crucial role in generating stable and natural free-form poses. To demonstrate the effectiveness of the proposed diversity-guided conditional vector $\mathbf{D_{\text{pose}}}$ and stability-guided conditional vector $\mathbf{S_{\text{pose}}}$, we evaluate the density metric on both static and dynamic audio, along with the absolute error at final convergence. As shown in Tab.~\ref{AudioPose ab}, incorporating $\mathbf{D_{\text{pose}}}$ and $\mathbf{S_{\text{pose}}}$ either individually or together consistently produces natural avatar. However, the absence of $\mathbf{S_{\text{pose}}}$ leads to unstable head generation with frequent jittering, resulting in an increased density metric. On the other hand, excluding $\mathbf{D_{\text{pose}}}$ results in larger absolute poses errors, and the diversity in silent segments becomes excessively high, making it difficult to achieve the desired small-scale head motion.

\subsubsection{Effectiveness of $\mathbf{D_{\text{exp}}}$ and $\mathbf{S_{\text{exp}}}$}

In the Audio-driven talking avatar, we evaluate the impact of the proposed $\mathbf{D_{\text{exp}}}$ and $\mathbf{S_{\text{exp}}}$ on facial expressions, specifically focusing on blinking behavior. The evaluation metric is the EAR (Eye Aspect Ratio) calculated for each frame, along with line plots of blendshapes for the left and right eyes predicted by the Auido2Pose model. As shown in Fig.\ref{fig:ab_EXP}, when $\mathbf{D_{\text{exp}}}$ or $\mathbf{S_{\text{exp}}}$ is not incorporated, the rendered image exhibits either a single blink over the 22-second driven audio or incomplete eye closures despite EAR fluctuations. Under the combined effect of $\mathbf{D_{\text{exp}}}$ and diversity-guided $\mathbf{S_{\text{exp}}}$, periodic fluctuations are guided by $\mathbf{D_{\text{exp}}}$, while $\mathbf{S_{\text{exp}}}$ accentuates each peak (as illustrated in the comparison between (b) and (c)), resulting in more natural and rhythmic blinking motion.

\begin{table}[h]
\centering

\begin{tabular}{lccc}
\hline
\multirow{2}{*}{Method}          & \textbf{L1-error} $\downarrow$     & \textbf{Diversity}  $\uparrow$   & \textbf{Diversity}$\downarrow$ \\
                                 & (\textbf{latest iter})     & (\textbf{Normal})    & (\textbf{Silence})        \\ \hline
w/o $\mathbf{D}_{\text{pose}}$ ,$\mathbf{S}_{\text{pose}}$                     &1.4910     &None     &None    \\                                
w/o $\mathbf{D}_{\text{pose}}$                                                 &0.1677     &0.3333     &0.2142    \\

w/o $\mathbf{S}_{\text{pose}}$                                                 &0.009       &0.3067     &0.1584       \\                                               

Ours                                                                           & 0.009      & 0.3117               & 0.1209                     \\\hline
\end{tabular}

\caption{Ablation the diversity and L1 error of the proposed $\mathbf{D_{\text{pose}}}$ and $\mathbf{S_{\text{pose}}}$ in OOD audio inference. Each conditional template contribute largely to generate realistic head motion.}
\label{AudioPose ab}
\end{table}

\section{Conclusion}
In this paper, we propose SyncAnimation, which aims to achieve three key objectives for modern talking avatars: generalized audio-to-head poses matching, consistent audio-to-facial expression synchronization, and jointly generative upper-body avatars. Our framework integrates AudioPose Syncer, AudioEmotion Syncer, and High-Synchronization Human Renderer modules. These components enable the generation of stable synchronized poses, upper-body movement, and realistic facial expressions while preserving subject identity, even from monocular or noisy inputs. Extensive experiments demonstrate that our method achieves the three goals of modern talking avatars and outperforms existing methods. Furthermore, the SyncAnimation is the first NeRF-based framework capable of generating realistic upper-body avatars with the same size as the original video, driven by audio using monocular inputs, while meeting real-time requirements.
%
%
\bibliographystyle{named}
\bibliography{ijcai25}

\begin{thebibliography}{}

\bibitem[\protect\citeauthoryear{Baevski \bgroup \em et al.\egroup }{2020}]{wav2vec}
Alexei Baevski, Henry Zhou, Abdelrahman Mohamed, and Michael Auli.
\newblock wav2vec 2.0: a framework for self-supervised learning of speech representations.
\newblock In {\em Proceedings of the 34th International Conference on Neural Information Processing Systems}, NIPS '20, Red Hook, NY, USA, 2020. Curran Associates Inc.

\bibitem[\protect\citeauthoryear{Baltru{\v{s}}aitis \bgroup \em et al.\egroup }{2015}]{baltruvsaitis2015cross}
Tadas Baltru{\v{s}}aitis, Marwa Mahmoud, and Peter Robinson.
\newblock Cross-dataset learning and person-specific normalisation for automatic action unit detection.
\newblock In {\em 2015 11th IEEE international conference and workshops on automatic face and gesture recognition (FG)}, volume~6, pages 1--6. IEEE, 2015.

\bibitem[\protect\citeauthoryear{Chen \bgroup \em et al.\egroup }{2018}]{chen2018lip}
Lele Chen, Zhiheng Li, Ross~K Maddox, Zhiyao Duan, and Chenliang Xu.
\newblock Lip movements generation at a glance.
\newblock In {\em Proceedings of the European conference on computer vision (ECCV)}, pages 520--535, 2018.

\bibitem[\protect\citeauthoryear{Chen \bgroup \em et al.\egroup }{2024}]{chen2024echomimic}
Zhiyuan Chen, Jiajiong Cao, Zhiquan Chen, Yuming Li, and Chenguang Ma.
\newblock Echomimic: Lifelike audio-driven portrait animations through editable landmark conditions.
\newblock {\em arXiv preprint arXiv:2407.08136}, 2024.

\bibitem[\protect\citeauthoryear{Cheng \bgroup \em et al.\egroup }{2022}]{cheng2022videoretalking}
Kun Cheng, Xiaodong Cun, Yong Zhang, Menghan Xia, Fei Yin, Mingrui Zhu, Xuan Wang, Jue Wang, and Nannan Wang.
\newblock Videoretalking: Audio-based lip synchronization for talking head video editing in the wild.
\newblock In {\em SIGGRAPH Asia 2022 Conference Papers}, pages 1--9, 2022.

\bibitem[\protect\citeauthoryear{Geirhos \bgroup \em et al.\egroup }{2020}]{geirhos2020shortcut}
Robert Geirhos, J{\"o}rn-Henrik Jacobsen, Claudio Michaelis, Richard Zemel, Wieland Brendel, Matthias Bethge, and Felix~A Wichmann.
\newblock Shortcut learning in deep neural networks.
\newblock {\em Nature Machine Intelligence}, 2(11):665--673, 2020.

\bibitem[\protect\citeauthoryear{Guo \bgroup \em et al.\egroup }{2021}]{guo2021ad}
Yudong Guo, Keyu Chen, Sen Liang, Yong-Jin Liu, Hujun Bao, and Juyong Zhang.
\newblock Ad-nerf: Audio driven neural radiance fields for talking head synthesis.
\newblock In {\em Proceedings of the IEEE/CVF international conference on computer vision}, pages 5784--5794, 2021.

\bibitem[\protect\citeauthoryear{Guo \bgroup \em et al.\egroup }{2024}]{guo2024i2v}
Xun Guo, Mingwu Zheng, Liang Hou, Yuan Gao, Yufan Deng, Pengfei Wan, Di~Zhang, Yufan Liu, Weiming Hu, Zhengjun Zha, et~al.
\newblock I2v-adapter: A general image-to-video adapter for diffusion models.
\newblock In {\em ACM SIGGRAPH 2024 Conference Papers}, pages 1--12, 2024.

\bibitem[\protect\citeauthoryear{Hannun \bgroup \em et al.\egroup }{2014}]{hannun2014deep}
Awni~Y. Hannun, Carl Case, Jared Casper, Bryan Catanzaro, Gregory~Frederick Diamos, Erich Elsen, Ryan~J. Prenger, Sanjeev Satheesh, Shubho Sengupta, Vinay Rao, Adam Coates, and A.~Ng.
\newblock Deep speech: Scaling up end-to-end speech recognition.
\newblock {\em arXiv preprint arXiv:1412.5567}, 2014.

\bibitem[\protect\citeauthoryear{Haque and Yumak}{2023}]{FaceXHubert}
Kazi~Injamamul Haque and Zerrin Yumak.
\newblock Facexhubert: Text-less speech-driven e(x)pressive 3d facial animation synthesis using self-supervised speech representation learning.
\newblock In {\em Proceedings of the 25th International Conference on Multimodal Interaction}, ICMI '23, page 282–291. Association for Computing Machinery, 2023.

\bibitem[\protect\citeauthoryear{Heusel \bgroup \em et al.\egroup }{2017}]{heusel2017gans}
Martin Heusel, Hubert Ramsauer, Thomas Unterthiner, Bernhard Nessler, and Sepp Hochreiter.
\newblock Gans trained by a two time-scale update rule converge to a local nash equilibrium.
\newblock {\em Advances in neural information processing systems}, 30, 2017.

\bibitem[\protect\citeauthoryear{Hore and Ziou}{2010}]{hore2010image}
Alain Hore and Djemel Ziou.
\newblock Image quality metrics: Psnr vs. ssim.
\newblock In {\em 2010 20th international conference on pattern recognition}, pages 2366--2369. IEEE, 2010.

\bibitem[\protect\citeauthoryear{Hsu \bgroup \em et al.\egroup }{2021}]{Hubert}
Wei-Ning Hsu, Benjamin Bolte, Yao-Hung~Hubert Tsai, Kushal Lakhotia, Ruslan Salakhutdinov, and Abdelrahman Mohamed.
\newblock Hubert: Self-supervised speech representation learning by masked prediction of hidden units.
\newblock {\em IEEE/ACM Trans. Audio, Speech and Lang. Proc.}, page 3451–3460, October 2021.

\bibitem[\protect\citeauthoryear{Hu}{2024}]{hu2024animate}
Li~Hu.
\newblock Animate anyone: Consistent and controllable image-to-video synthesis for character animation.
\newblock In {\em Proceedings of the IEEE/CVF Conference on Computer Vision and Pattern Recognition}, pages 8153--8163, 2024.

\bibitem[\protect\citeauthoryear{Kim \bgroup \em et al.\egroup }{2024}]{kim2024nerffacespeech}
Gihoon Kim, Kwanggyoon Seo, Sihun Cha, and Junyong Noh.
\newblock Nerffacespeech: One-shot audio-diven 3d talking head synthesis via generative prior.
\newblock {\em arXiv preprint arXiv:2405.05749}, 2024.

\bibitem[\protect\citeauthoryear{Li \bgroup \em et al.\egroup }{2023}]{li2023efficient}
Jiahe Li, Jiawei Zhang, Xiao Bai, Jun Zhou, and Lin Gu.
\newblock Efficient region-aware neural radiance fields for high-fidelity talking portrait synthesis.
\newblock In {\em Proceedings of the IEEE/CVF International Conference on Computer Vision}, pages 7568--7578, 2023.

\bibitem[\protect\citeauthoryear{Mildenhall \bgroup \em et al.\egroup }{2021}]{mildenhall2021nerf}
Ben Mildenhall, Pratul~P Srinivasan, Matthew Tancik, Jonathan~T Barron, Ravi Ramamoorthi, and Ren Ng.
\newblock Nerf: Representing scenes as neural radiance fields for view synthesis.
\newblock {\em Communications of the ACM}, 65(1):99--106, 2021.

\bibitem[\protect\citeauthoryear{M\"{u}ller \bgroup \em et al.\egroup }{2022}]{hashencoding}
Thomas M\"{u}ller, Alex Evans, Christoph Schied, and Alexander Keller.
\newblock Instant neural graphics primitives with a multiresolution hash encoding.
\newblock {\em ACM Trans. Graph.}, 41(4), July 2022.

\bibitem[\protect\citeauthoryear{Peng \bgroup \em et al.\egroup }{2023}]{peng2023emotalk}
Ziqiao Peng, Haoyu Wu, Zhenbo Song, Hao Xu, Xiangyu Zhu, Jun He, Hongyan Liu, and Zhaoxin Fan.
\newblock Emotalk: Speech-driven emotional disentanglement for 3d face animation.
\newblock In {\em Proceedings of the IEEE/CVF International Conference on Computer Vision}, pages 20687--20697, 2023.

\bibitem[\protect\citeauthoryear{Peng \bgroup \em et al.\egroup }{2024}]{peng2024synctalk}
Ziqiao Peng, Wentao Hu, Yue Shi, Xiangyu Zhu, Xiaomei Zhang, Hao Zhao, Jun He, Hongyan Liu, and Zhaoxin Fan.
\newblock Synctalk: The devil is in the synchronization for talking head synthesis.
\newblock In {\em Proceedings of the IEEE/CVF Conference on Computer Vision and Pattern Recognition}, pages 666--676, 2024.

\bibitem[\protect\citeauthoryear{Prajwal \bgroup \em et al.\egroup }{2020}]{prajwal2020lip}
KR~Prajwal, Rudrabha Mukhopadhyay, Vinay~P Namboodiri, and CV~Jawahar.
\newblock A lip sync expert is all you need for speech to lip generation in the wild.
\newblock In {\em Proceedings of the 28th ACM international conference on multimedia}, pages 484--492, 2020.

\bibitem[\protect\citeauthoryear{Qian \bgroup \em et al.\egroup }{2021}]{qian2021speech}
Shenhan Qian, Zhi Tu, Yihao Zhi, Wen Liu, and Shenghua Gao.
\newblock Speech drives templates: Co-speech gesture synthesis with learned templates.
\newblock In {\em Proceedings of the IEEE/CVF International Conference on Computer Vision}, pages 11077--11086, 2021.

\bibitem[\protect\citeauthoryear{Ruiz \bgroup \em et al.\egroup }{2018}]{ruiz2018fine}
Nataniel Ruiz, Eunji Chong, and James~M Rehg.
\newblock Fine-grained head pose estimation without keypoints.
\newblock In {\em Proceedings of the IEEE conference on computer vision and pattern recognition workshops}, pages 2074--2083, 2018.

\bibitem[\protect\citeauthoryear{Soukupova and Cech}{2016}]{soukupova2016eye}
Tereza Soukupova and Jan Cech.
\newblock Eye blink detection using facial landmarks.
\newblock In {\em 21st computer vision winter workshop, Rimske Toplice, Slovenia}, volume~2, 2016.

\bibitem[\protect\citeauthoryear{Tan \bgroup \em et al.\egroup }{2025}]{tan2025edtalk}
Shuai Tan, Bin Ji, Mengxiao Bi, and Ye~Pan.
\newblock Edtalk: Efficient disentanglement for emotional talking head synthesis.
\newblock In {\em European Conference on Computer Vision}, pages 398--416. Springer, 2025.

\bibitem[\protect\citeauthoryear{Tian \bgroup \em et al.\egroup }{2025}]{tian2025emo}
Linrui Tian, Qi~Wang, Bang Zhang, and Liefeng Bo.
\newblock Emo: Emote portrait alive generating expressive portrait videos with audio2video diffusion model under weak conditions.
\newblock In {\em European Conference on Computer Vision}, pages 244--260. Springer, 2025.

\bibitem[\protect\citeauthoryear{Wang \bgroup \em et al.\egroup }{2004}]{wang2004image}
Zhou Wang, Alan~C Bovik, Hamid~R Sheikh, and Eero~P Simoncelli.
\newblock Image quality assessment: from error visibility to structural similarity.
\newblock {\em IEEE transactions on image processing}, 13(4):600--612, 2004.

\bibitem[\protect\citeauthoryear{Wang \bgroup \em et al.\egroup }{2024}]{wang2024v}
Cong Wang, Kuan Tian, Jun Zhang, Yonghang Guan, Feng Luo, Fei Shen, Zhiwei Jiang, Qing Gu, Xiao Han, and Wei Yang.
\newblock V-express: Conditional dropout for progressive training of portrait video generation.
\newblock {\em arXiv preprint arXiv:2406.02511}, 2024.

\bibitem[\protect\citeauthoryear{Xie \bgroup \em et al.\egroup }{2024}]{xie2024x}
You Xie, Hongyi Xu, Guoxian Song, Chao Wang, Yichun Shi, and Linjie Luo.
\newblock X-portrait: Expressive portrait animation with hierarchical motion attention.
\newblock In {\em ACM SIGGRAPH 2024 Conference Papers}, pages 1--11, 2024.

\bibitem[\protect\citeauthoryear{Xu \bgroup \em et al.\egroup }{2024}]{xu2024hallo}
Mingwang Xu, Hui Li, Qingkun Su, Hanlin Shang, Liwei Zhang, Ce~Liu, Jingdong Wang, Yao Yao, and Siyu Zhu.
\newblock Hallo: Hierarchical audio-driven visual synthesis for portrait image animation.
\newblock {\em arXiv preprint arXiv:2406.08801}, 2024.

\bibitem[\protect\citeauthoryear{Ye \bgroup \em et al.\egroup }{2023a}]{ye2023geneface++}
Zhenhui Ye, Jinzheng He, Ziyue Jiang, Rongjie Huang, Jiawei Huang, Jinglin Liu, Yi~Ren, Xiang Yin, Zejun Ma, and Zhou Zhao.
\newblock Geneface++: Generalized and stable real-time audio-driven 3d talking face generation.
\newblock {\em arXiv preprint arXiv:2305.00787}, 2023.

\bibitem[\protect\citeauthoryear{Ye \bgroup \em et al.\egroup }{2023b}]{ye2023geneface}
Zhenhui Ye, Ziyue Jiang, Yi~Ren, Jinglin Liu, Jinzheng He, and Zhou Zhao.
\newblock Geneface: Generalized and high-fidelity audio-driven 3d talking face synthesis.
\newblock {\em arXiv preprint arXiv:2301.13430}, 2023.

\bibitem[\protect\citeauthoryear{Zhang \bgroup \em et al.\egroup }{2018}]{zhang2018unreasonable}
Richard Zhang, Phillip Isola, Alexei~A Efros, Eli Shechtman, and Oliver Wang.
\newblock The unreasonable effectiveness of deep features as a perceptual metric.
\newblock In {\em Proceedings of the IEEE conference on computer vision and pattern recognition}, pages 586--595, 2018.

\bibitem[\protect\citeauthoryear{Zhang \bgroup \em et al.\egroup }{2021}]{zhang2021flow}
Zhimeng Zhang, Lincheng Li, Yu~Ding, and Changjie Fan.
\newblock Flow-guided one-shot talking face generation with a high-resolution audio-visual dataset.
\newblock In {\em Proceedings of the IEEE/CVF Conference on Computer Vision and Pattern Recognition}, pages 3661--3670, 2021.

\bibitem[\protect\citeauthoryear{Zhang \bgroup \em et al.\egroup }{2023a}]{zhang2023sadtalker}
Wenxuan Zhang, Xiaodong Cun, Xuan Wang, Yong Zhang, Xi~Shen, Yu~Guo, Ying Shan, and Fei Wang.
\newblock Sadtalker: Learning realistic 3d motion coefficients for stylized audio-driven single image talking face animation.
\newblock In {\em Proceedings of the IEEE/CVF Conference on Computer Vision and Pattern Recognition}, pages 8652--8661, 2023.

\bibitem[\protect\citeauthoryear{Zhang \bgroup \em et al.\egroup }{2023b}]{zhang2023dinet}
Zhimeng Zhang, Zhipeng Hu, Wenjin Deng, Changjie Fan, Tangjie Lv, and Yu~Ding.
\newblock Dinet: Deformation inpainting network for realistic face visually dubbing on high resolution video.
\newblock In {\em Proceedings of the AAAI Conference on Artificial Intelligence}, volume~37, pages 3543--3551, 2023.

\bibitem[\protect\citeauthoryear{Zhong \bgroup \em et al.\egroup }{2023}]{zhong2023identity}
Weizhi Zhong, Chaowei Fang, Yinqi Cai, Pengxu Wei, Gangming Zhao, Liang Lin, and Guanbin Li.
\newblock Identity-preserving talking face generation with landmark and appearance priors.
\newblock In {\em Proceedings of the IEEE/CVF Conference on Computer Vision and Pattern Recognition}, pages 9729--9738, 2023.

\end{thebibliography}
\end{document}